\documentclass{article} % For LaTeX2e
\usepackage{iclr2026_delta,times}
\usepackage{float}
\usepackage[utf8]{inputenc}
\usepackage{amsmath}
\usepackage{algorithm}

\usepackage{algpseudocode}
\usepackage{booktabs} % For professional looking tables
\usepackage{soul}
\usepackage{url}
\usepackage{graphicx}
\usepackage{amsthm}
\usepackage{amsfonts}       % blackboard math symbols
\usepackage{nicefrac}       % compact symbols for 1/2, etc.
\usepackage{microtype}      % microtypography
\usepackage{xcolor}         % colors
\usepackage{paralist}
\usepackage{xspace}
\usepackage{amsmath,amsfonts,bm}

\def\eqref#1{equation~\ref{#1}}
\def\1{\bm{1}}

\DeclareMathAlphabet{\mathsfit}{\encodingdefault}{\sfdefault}{m}{sl}
\SetMathAlphabet{\mathsfit}{bold}{\encodingdefault}{\sfdefault}{bx}{n}

\def\gL{{\mathcal{L}}}

\def\bfx{\mathbf{x}}

\def\bfI{\mathbf{I}}

\usepackage{hyperref}
\usepackage{url}

\title{B-DENSE: Branching for Dense Ensemble Network Supervision Efficiency}

\author{Cherish Puniani, Tushar Kumar, Arnav Bendre, Gaurav Kumar, Shree Singhi \\
\vspace{2mm} % Adds a small gap between names and institute
\textit{Indian Institute of Technology, Roorkee} \\
Roorkee, Uttarakhand, India \\
\vspace{2mm} % Adds a small gap before emails
\texttt{cherish\_p@me.iitr.ac.in, tushar\_k@cy.iitr.ac.in,} \\
\texttt{arnav\_b@hre.iitr.ac.in, gaurav\_k@mfs.iitr.ac.in,} \\
\texttt{shree\_s@mfs.iitr.ac.in}
}
\iclrfinalcopy 
\begin{document}

\maketitle

\begin{abstract}
Inspired by non-equilibrium thermodynamics, diffusion models have achieved state-of-the-art performance in generative modeling. However, their iterative sampling nature results in high inference latency, which is typically required to maintain image quality. While recent efforts in distillation techniques have improved sample quality with fewer steps, they discard intermediate trajectory steps. By discarding intermediate trajectory steps, these methods lose structural information, resulting in significant discretization errors. To mitigate this issue, we propose a novel framework, \textbf{B-DENSE}, that leverages multi-branch trajectory alignment. We train the  student model using branches that simultaneously map to the entire sequence of the teacher's target timesteps.
We modify the student architecture to output K$-$fold expanded channels. Each channel subset corresponds to a specific branch representing a discrete intermediate step in the teacher’s trajectory. By enforcing intermediate trajectory alignment, the student model learns to navigate the solution space from the earliest stages of training, leading to better image generation quality than the baseline distillation frameworks.
\end{abstract}

\section{Introduction}
Diffusion models \cite{ho2020ddpm,sohl2015deep}have recently come to dominate image synthesis, surpassing previous paradigms such as Generative Adversarial Networks\cite{heusel2017gans} and Variational Autoencoders \cite{kingma2013auto}. This is largely attributed to a stable training objective and robust framework that learns to reverse a gradual noise injection process by iterative sampling. By transforming a standard normal prior into high-fidelity data through a sequence of learned Gaussian transitions, these models achieve exceptional sample quality and mode coverage. However, these advantages come at a high computational cost; achieving high-quality results typically requires hundreds or thousands of iterative denoising steps, rendering inference slow and compute-intensive. Numerous studies focus on mitigating this bottleneck by either designing faster ODE solvers that speed up the inference process without changing the diffusion model's parameters (DDIM, Euler, Heun, DPM-Solver++), or by distilling slow but high-quality models into faster models that take far fewer steps. Many distillation methods (\cite{salimans2022progressive,zhou2024simple,kim2023consistency} \cite{kim2024consistencytrajectorymodelslearning} \cite{Xiang_2025_CVPR})supervise the student only at a sparse set of timestamps,
matching the teacher at the endpoints of a collapsed interval. This sparse supervision discards critical trajectory geometry, leading to significant discretization errors and increased sensitivity to step reductions.

We propose B-DENSE, a novel distillation framework that addresses these limitations by modifying the architecture to explicitly align the student with the teacher’s entire denoising trajectory over each collapsed interval without any significant
computation overhead. Our method modifies the student architecture to output $K\cdot C$ channels, organized into $K$ parallel branches, where each branch predicts the teacher’s denoised state at a distinct intermediate timestamp within a teacher interval. During training, the teacher generates the full sequence of intermediate states, and the student is supervised with a multi-branch loss that enforces agreement at all intermediate points, rather than only at the interval endpoints. B-DENSE recovers the fine-grained teacher updates typically discarded during distillation. By treating these branches as an ensemble of local predictors, the model approximates the full trajectory more accurately. This dense alignment minimizes discretization errors.

Across experiments, B-DENSE demonstrates better FID scores, with notable improvements in low-step sampling regimes. Importantly, the method incurs minimal additional overhead; the multi-branch output requires $(K-1)$ additional channel dimensions, while teacher target generation remains the dominant cost, making the approach broadly compatible with existing U-Net based diffusion architectures. B-DENSE provides a computationally efficient mechanism for dense trajectory supervision in diffusion distillation. We provide a theoretical interpretation of the method as a piecewise quadrature approximation of the probability-flow ODE, which helps explain its reduced discretization errors and better generation quality when aggressively reducing the number of sampling steps. Through empirical evaluation, we found that B-DENSE achieves better sample quality than established baselines, particularly in low-step regimes, while maintaining the same inference-time complexity.

\begin{figure}
    \centering
    \includegraphics[width=0.8\textwidth]{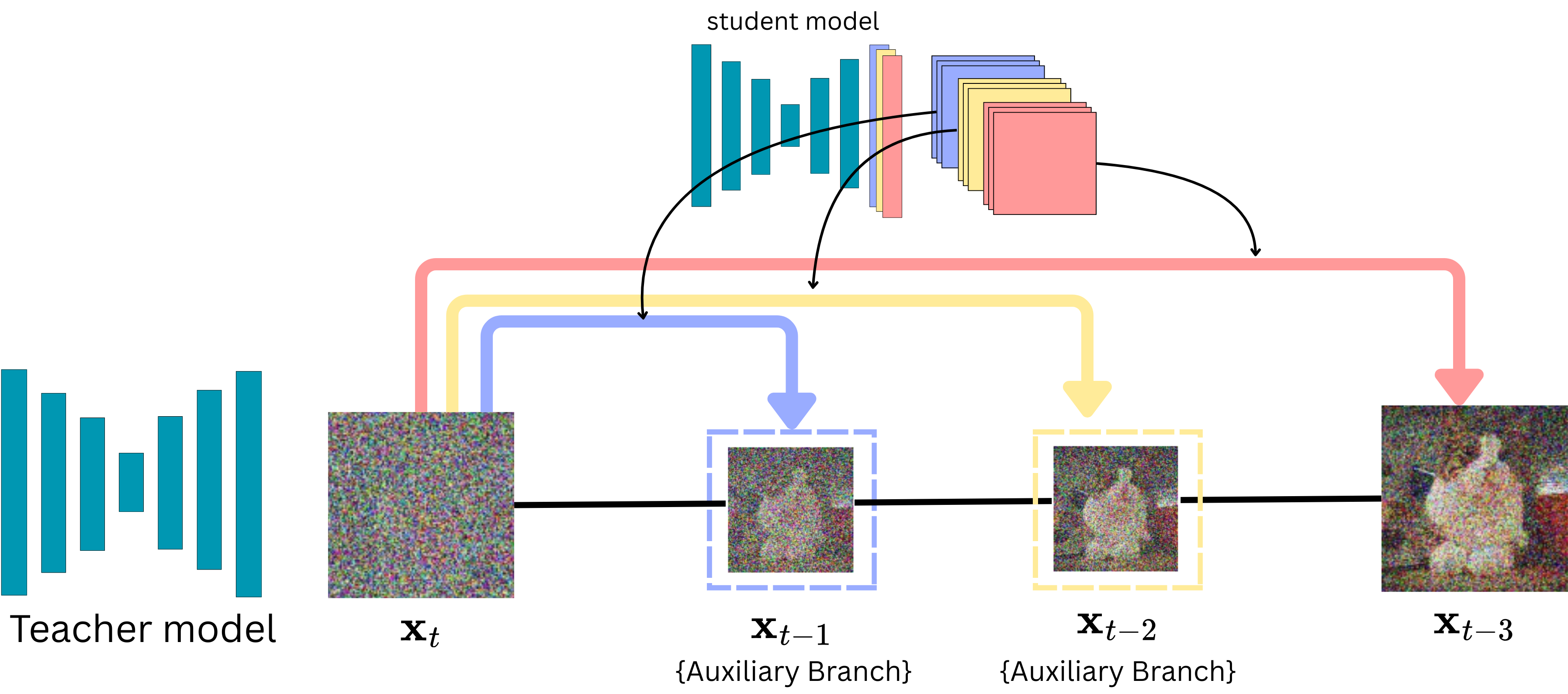}
    \caption{This visualization illustrates a single iteration of the proposed B-DENSE method. Here, $x_t$ represents a noisy image at a random time step $t$. At the same time, $x_{t-3}$ is output directly mapped by the standard distillation method, and  $x_{t-1}$ and $x_{t-2}$ are consecutive intermediate denoised outputs that will be skipped by the standard distillation method. The blue, yellow, and red channels indicate branched channels for $K=3$. Blue, yellow, and red arrows illustrate the mapping of each three-channel configuration to the final output and to an intermediate step used to calculate the reconstruction loss for each corresponding branch.}
    \label{fig:distillation_step}
\end{figure}
\section{Background}
\subsection{Diffusion framework}
Diffusion models are generative models inspired by non-equilibrium statistical physics \cite{sohl2015deep}. They progressively inject Gaussian noise into the data, destroying its structure through an iterative forward diffusion process. The models then learn a reverse process that reconstructs the original image from pure noise. Theoretically, these models can be described using stochastic differential equations (SDEs), where the forward process is defined as:
$$dx = f(x, t)dt + g(t)dw_t$$
In this context, $f(\cdot, t)$ and $g(\cdot)$ denote the drift and diffusion coefficients, respectively, while $w_{t}$ represents the Wiener process. The generative model approximates the intractable reverse process using a reverse-time SDE to sample data from noise, defined as follows:
$$dx = [f(x, t) - g^2(t) \nabla_x \log p_t(x)]dt + g(t)d\bar{w}_t$$

Here $\nabla_x \log p_t(x)$ is known as the score \cite{song2019ncsn}, which represents the gradient of the log-density of the noisy data at each timestep $t$

To implement the reverse process, we parameterize a neural network $\epsilon_{\theta}(x, t)$ referred to as the noise-prediction model to estimate the score function. The model is trained by optimizing a variational lower bound, which simplifies to a weighted mean squared error (MSE) between the actual noise $\epsilon$ and the predicted noise:
$$L_{simple} = E_{x_0, \epsilon, t} [\|\epsilon - \epsilon_\theta(x_t, t)\|^2]$$

While the SDE framework is mathematically elegant, it presents several significant issues, such as slow sampling speed and stochasticity, because it requires many small discretization steps. Every time we sample from the same starting noise $x_T$, we get a different result. While diversity is usually beneficial, it makes precise latent-space manipulations impossible. To eliminate this stochasticity, the diffusion process can be reformulated as a deterministic Probability Flow Ordinary Differential Equation (ODE) \cite{song2021scorebasedgenerativemodelingstochastic}, also often called probability flow ODE, is defined as
$$dx = \left[ f(x, t) - \frac{1}{2}g(t)^2 \nabla_x \log p_t(x) \right] dt$$
The probability flow ODE transforms the stochastic process into a deterministic mapping. This ensures that a specific initial noise vector $x_T$ always collapses to the same image $x_0$. Consequently, the generative trajectory can be subsampled, enabling high-fidelity sample synthesis in as few as 10–50 steps rather than the canonical 1000.

\subsection{Distillation and Related Work}

\textbf{Samplers} are numerical algorithms designed to compute the reverse-time generation process, typically by formulating it as a PF-ODE that maps noise back to data. Because the standard denoising process is sequential and requires many iterations to resolve the continuous trajectory, solvers help speed up diffusion by employing more efficient discretization techniques (such as higher-order approximations like Heun or DPM-Solver) to estimate the correct path with fewer steps. This more efficient discretization reduces the NFE required to generate a high-quality image, thereby lowering the sampling latency without the need to retrain the underlying model.

\textbf{Distillation of diffusion models} is a training approach aimed at reducing the high computational costs associated with the standard iterative sampling process. This process usually involves hundreds or thousands of steps to generate a single image. In this framework, a 'student' model is trained to replicate the behavior of a pre-trained 'teacher' model but is optimized to follow the denoising trajectory using significantly fewer timesteps.

Progressive Distillation \cite{salimans2022progressivedistillationfastsampling} accelerates diffusion model sampling by iteratively training a student model to compress two consecutive deterministic steps of a teacher model, typically using a DDIM sampler, into a single step. To ensure stability as the number of steps decreases, the student model is often reparameterized to predict the clean data $x$ or the velocity $v$ directly rather than the noise $\epsilon$, and is trained using a Truncated SNR loss that maintains training signals even at zero signal-to-noise ratios. The student learns to predict a precise target value derived by analytically inverting the diffusion process to match the teacher's trajectory, allowing the sampling process to be recursively reduced from thousands of steps to as few as four with minimal loss in generation quality.

Simple and Fast Distillation (SFD) \cite{zhou2024simple} streamlines diffusion model acceleration by addressing the 'step mismatch' inefficiency, grounded in the theoretical observation that optimizing the gradient field at specific timestamps smoothly enhances the score function at neighbouring times, rendering dense fine-tuning unnecessary. Unlike standard 'local' distillation methods that minimize error at each step independently, SFD adopts a 'global' trajectory-matching framework in which the student model generates and imitates the complete sampling trajectory of a high-order teacher (e.g., DPM-Solver++), enabling it to correct accumulated discretization errors dynamically. By restricting optimization to the timestamps used during inference and using a simplified L1 loss, SFD eliminates the need for complex objectives while enabling variable-step sampling via step-conditioning .

Consistency Models \cite{song2023consistencymodels} accelerate the generative process by learning a function that maps any point along the probability flow ODE trajectory directly to its original clean data, enabling single-step generation. Unlike standard diffusion models that require iterative denoising, this approach enforces a self-consistency property, ensuring that widely separated points on the same trajectory are mapped to the exact same initial data point $x_0$. This formulation allows the model to generate high-quality samples from pure noise in a single forward pass, while also supporting multi-step sampling that trades increased computation for improved sample fidelity.

\begin{minipage}[t]{0.48\textwidth}
\vspace{0pt} % Forces the baseline to the top
\begin{algorithm}[H]
\caption{PD}
\label{pd}
\begin{algorithmic}
\State \textbf{Require:} Trained teacher model $\hat{x}_\eta(\mathbf{z}_t)$
\State \textbf{Require:} Data set $\mathcal{D}$
\State \textbf{Require:} Loss weight function $w()$
\State \textbf{Require:} Teacher steps to approximate $N$
\State
\For{$K$ iterations}
    \State $\phi \leftarrow \eta$ \hfill $\triangleright$ Init student from teacher
    \While{not converged}
        \State $\mathbf{x} \sim \mathcal{D}$
        \State $\epsilon \sim \mathcal{N}(0, I)$
        \State $t \sim [1, T]$
        \State $\mathbf{z}_t = \alpha_t \mathbf{x} + \sigma_t \epsilon$
        \State \# $2$ steps of DDIM with teacher
        \State $\hat{x}_{t-2}$ $\rightarrow $
        \text{latent } 
        \State $\lambda_t = \log[\alpha_t^2 / \sigma_t^2]$ 
        \State $\gL_\Phi =  w(\lambda_t) \|\hat{x}_{t-2} - \tilde{x}_\phi(\mathbf{z}_t)\|_2^2$
        \State $\phi \leftarrow \phi - \gamma \nabla_\phi \gL_\phi$
    \EndWhile
    \State $\eta \leftarrow \phi$ \hfill $\triangleright$ Student becomes teacher
    \State $T \leftarrow T/N$
\EndFor
\end{algorithmic}
\end{algorithm}
\end{minipage}
\hfill
% --- Algorithm 2: B-DENSE ---
\begin{minipage}[t]{0.48\textwidth}
\vspace{0pt} % Forces the baseline to the top
\begin{algorithm}[H]
\caption{PD + B-DENSE}
\label{pd_bdense}
\begin{algorithmic}
\State \textbf{Require:} Trained teacher model $\hat{x}_\eta(\mathbf{z}_t)$
\State \textbf{Require:} Data set $\mathcal{D}$
\State \textbf{Require:} Loss weight function $w()$
\State \textbf{Require:} Teacher steps to approximate $N$
\State \textbf{Require:} Weights $\{\lambda_k\}_{k=0}^{K-1}$
\For{$K$ iterations}
    \State $\phi \leftarrow \eta$ \hfill $\triangleright$ Init student from teacher
    \While{not converged}
        \State $\mathbf{x} \sim \mathcal{D}$
        \State $\epsilon \sim \mathcal{N}(0, I)$
        \State $t \sim [1, T]$
        \State $\mathbf{z}_t = \alpha_t \mathbf{x} + \sigma_t \epsilon$
        \State \# $K$ steps of DDIM with teacher
        \State $\text{Latents} \rightarrow [\hat{x}_{t}, \dots, \hat{x}_{t-K-1}] $
        \State $\text{Student outputs} \rightarrow [\tilde{x}_{t}, \dots, \tilde{x}_{t-K-1}]$
        
        \State $\gL_\phi =   \sum_{k=0}^{K-1}  \lambda_k \cdot \|{\hat{x}}_{t-k} - \tilde{x}_{t-k}(\mathbf{z}_{t})\|_2^2$
        \State $\phi \leftarrow \phi - \gamma \nabla_\phi \gL_\phi$
    \EndWhile
    \State $\eta \leftarrow \phi$ \hfill $\triangleright$ Student becomes teacher
    \State $T \leftarrow T/N$
\EndFor
\end{algorithmic}
\end{algorithm}
\end{minipage}
\vspace{1cm}

\section{Methodology}

\subsection{B-DENSE}
Most Diffusion Distillation frameworks only train on a subset of the teacher's outputs and discard intermediate trajectory steps. Consequently, these frameworks miss valuable supervision signals, leading to suboptimal convergence. B-DENSE leverages these intermediate teacher outputs, which are already available, to provide dense supervision to the student model. This reduces discretization errors and improves the quality of the final image.

B-DENSE can be interpreted as a more accurate instance of numerical integration over the generative trajectory defined by the teacher diffusion model. Instead of viewing denoising steps as discrete transitions, we treat them as evaluations of a continuous vector field $f(x, t)$ derived from the reverse-time probability flow ODE. By using the $K$ sub-intervals of the teacher’s decomposed sampling interval $[t_{n+1}, t_n]$  and training the student to match the teacher's predictions at each sub-interval boundary, the student approximates the total integral of $f$ via a structured piecewise quadrature. 

This method is a foundational architectural design that can be integrated into various distillation pipelines to ensure the student model navigates the solution space effectively from the earliest stages of training. We begin B-DENSE by using a pretrained diffusion model as the teacher rather than training it from scratch. Then, the student model is initialized as a copy of the teacher, with an important architectural change: to approximate $K$ teacher steps, the student’s final layer weights are initialized by repeating the teacher's weights $K$ times, to generate $K \cdot C$ channels instead of $C$, where $C$ represents the original channel dimension. The output tensor is partitioned into $K$ successive groups, each serving as an auxiliary ``branch" that predicts a specific intermediate state.

As in standard training, we sample data and add noise; however, the key difference lies in the training target, which is obtained by running the teacher model for $K$ inference steps and storing the full sequence of intermediate denoised images. To modulate the supervision signal strength across the trajectory, we assign weights $\{\lambda_i\}_{i=0}^{k-1}$ to the reconstruction loss of each corresponding branch, i.e. the new combined objective loss becomes $ \lambda_0\gL(x_0, \hat{x}_0) + \lambda_1\gL(x_1, \hat{x}_1)  +\cdots + \lambda_{k-1}\gL(x_{k-1}, \hat{x}_{k-1})$ where $\gL$ is the objective loss of the underlying distillation algorithm, $x_i$ is the $i^{th}$ target output generated by teacher and $\hat{x}_i$ is the $i^{th}$ student prediction. A comprehensive derivation regarding the selection of these weights is provided in the Appendix \autoref{app:optimal_weights}. Inference is performed using the weights corresponding to the final channel dimensions, which learns to map to the teacher’s interval endpoint.

Since the primary bottlenecks, teacher target generation and student gradient computation, remain unchanged, the multi-branch expansion is nearly cost-free. The architectural expansion to a multi-channel output ($K \cdot C$) is confined to the final layer, requiring only $K-1$ additional convolutional filters. This introduces a negligible computational overhead, increasing total FLOPs in the order of $\sim 0.01\%$ relative to the standard U-Net or Transformer backbone. Consequently, B-DENSE offers a high-efficiency pathway to increase training efficiency with minimal cost, making it broadly applicable across diverse distillation frameworks. We apply the B-DENSE framework on the Progressive Distillation, as shown in Algorithms \autoref{pd} and \autoref{pd_bdense}, and the SFD algorithm, as shown in Algorithms \autoref{SFD} and \autoref{SFD_bdense}.

\subsection{Theoretical Analysis}
The effectiveness of B-DENSE can be formally justified by viewing the sampling process through the lens of the Probability Flow ODE.

\textbf{Theoretical Foundation}
 The generation process in diffusion models is fundamentally defined as solving the reverse-time Probability Flow ODE. Under this formulation, the data trajectory is governed by the following ODE:
\begin{equation}
    dx = \epsilon_\theta(x, t)dt
\end{equation}
where $x(t)$ represents the data state and $\epsilon_\theta$ denotes the learned vector field. To transition from a noisy state at time $t_{n+1}$ to a cleaner state at $t_n$, the ideal solver must compute the integral of this vector field:
\begin{equation}
    x(t_n) = x(t_{n+1}) + \int_{t_{n+1}}^{t_n} \epsilon_\theta(x(u), u) du
\end{equation}
The accuracy of the generative process depends entirely on the fidelity with which this integral is approximated during sampling.

\textbf{The Problem with Sparse Supervision (Baseline Distillation)}
Standard distillation techniques employ what we classify as Sparse Supervision. In these frameworks, the student model $S_\phi$ is trained to predict the integral's endpoint $x(t_{n-k})$ directly from the starting point $x(t_{n})$ in a single step, minimizing the loss:
\begin{equation}
\mathcal{L}_{\text{sparse}} 
= \left\| S_\phi\!\left({x_{t_n}}\right) - x_{t_{n-k}}\right \|^2
\end{equation}

This formulation treats the integral term $\int_{t_{n+1}}^{t_n} \epsilon_\theta du$ as a black box. The student attempts to learn the result of the integral without explicit constraints on the path taken.

As noted in SFD \cite{zhou2024simple}, diffusion trajectories often exhibit high curvature. In these regions, there exists a manifold of incorrect paths that $S_\phi$ could hallucinate. This lack of path constraints leads to suboptimal results and high discretization errors, as the student is prone to wandering off the true trajectory manifold during the early stages of training.

\noindent\textbf{B-DENSE as Dense Trajectory Supervision}
To address the limitations of sparse supervision, we introduce Dense Trajectory Supervision, motivated by the Ensemble Parallel Direction (EPD) solver \cite{zhu2025distillingparallelgradientsfast}. EPD solver demonstrates that utilizing intermediate gradients at times $\tau_k \in [t_{n+1}, t_n]$ significantly reduces the local truncation error of the integral approximation. While EPD calculates these intermediate values in parallel during inference to improve the integration step:
\begin{equation}
    x_{t_n} = x_{t_{n+1}} + h \sum_{k=1}^{K} \lambda_k \epsilon_\theta(x_{\tau_k}, \tau_k)
\end{equation}
B-DENSE forces the student to learn these states during training.

\noindent\textbf{Formulation:} We discretize the distillation interval $[t_{n+1}, t_n]$ into $K$ intermediate control points $\tau_0, \tau_1, \dots, \tau_{K-1}$, where $\tau_K = t_n$ which are teacher model's intermediate steps. The student model architecture is modified to output a set of predictions corresponding to these time steps:
\begin{equation}
    S_\phi(x_{t_{n+1}}) = \{ \hat{x}_{\tau_0}, \hat{x}_{\tau_2}, ..., \hat{x}_{\tau_{K-1}} \}
\end{equation}
This mapping is performed while the shared backbone is conditioned on $t_{n+1}$. By supervising each branch $k$ against its corresponding teacher target at $\tau_k$, the network is forced to internalize the local velocity of the probability-flow ODE, effectively learning the sub-integrals of the vector field.
Consequently, we define the Dense Loss Function as:
\begin{equation}
    \gL_{\text{branch}} = \sum_{k=0}^{K-1} w_k \cdot || \hat{x}_{\tau_k} - x_{\text{teacher}}(\tau_k) ||^2
\end{equation}
B-DENSE functions as a pinned numerical integrator. By enforcing the loss at each intermediate $\tau_k$, we constrain the student’s learned vector field to satisfy the ODE flow property at multiple points simultaneously
If we view the student's prediction for a branch $k$ as an approximation of the sub-integral:
\begin{equation}
    \hat{x}_{\tau_k} \approx x(t_{n+1}) + \int_{t_{n+1}}^{\tau_k} \epsilon_{\text{teacher}}(x(u), u) du
\end{equation}
Our method ensures that the student does not just learn the total integral, i.e, endpoint matching, but implicitly learns the sub-integrals. This constrains the parameter search space of $\phi$, effectively forcing the student to align with the teacher’s denoising trajectory.

% --- Algorithm 3: SFD ---
\begin{minipage}[t]{0.48\textwidth}
\vspace{0pt} 
\begin{algorithm}[H]
\caption{SFD}
\label{SFD}
\begin{algorithmic}
\State \textbf{Require:} Student $\bfx^\phi$, Teacher $\eta$
\State \textbf{Require:} ODE  ${\rm{Solver}}$
\State 
\While{not converged}
    \State Sample $\bfx_{n+1} \sim \mathcal{N}(\bfx_0;{t_{n+1}^2}\bfI)$
    \For{$n=N-1$ \textbf{to} $0$}
        \State $\bfx_n^\phi = {\rm{Euler}}(\bfx_{n+1}, t_{n+1}, t_n, 1; \phi)$
        \State $\tilde{\bfx}_{n} = {\rm{Solver}}(\bfx_{n+1}, t_{n+1}, t_n, K; \eta)_K$
        \State $\gL\phi = d(\bfx_n^\phi, \tilde{\bfx}_{n})$
        \State $\phi \leftarrow \phi - \eta\nabla_\phi \gL\phi$
        \State $\bfx_n = {\rm{detach}}(\bfx_n^\phi)$
    \EndFor
\EndWhile
\end{algorithmic}
\end{algorithm}
\end{minipage}
\hfill
% --- Algorithm 4: B-DENSE ---
\begin{minipage}[t]{0.48\textwidth}
\vspace{0pt}
\begin{algorithm}[H]
\caption{SFD + B-DENSE}
\label{SFD_bdense}
\begin{algorithmic} % [1] adds line numbers if desired
\State \textbf{Require:} Branched model $\phi_{aug}$, Teacher $\eta$
\State \textbf{Require:} Weights $\{\lambda_k\}_{k=0}^{K-1}$
\State 
\While{not converged}
    \State Sample $\mathbf{x}_{n+1} \sim \mathcal{N}(\mathbf{x}_0; t_{n+1}^2 \mathbf{I})$
    \For{$n = N - 1$ \textbf{to} $0$}
        \State $\mathbf{X}_n^\phi = \text{Euler}(\mathbf{x}_{n+1}, t_{n+1}, t_n, 1; \phi_{aug})$
        \State $\tilde{\mathbf{X}}_n = \text{Solver}(\mathbf{x}_{n+1}, t_{n+1}, t_n, K; \eta)$
        \State $\gL\phi = \sum_{k=0}^{K-1} \lambda_k \cdot d(\mathbf{X}_{n,k}^\phi, \tilde{\mathbf{X}}_{n,k})$
        \State $\phi_{aug} \leftarrow \phi_{aug} - \eta \nabla_{\phi_{aug}} \gL\phi$
        \State $\mathbf{x}_n = \text{detach}(\mathbf{X}_{n,K}^\phi)$
    \EndFor
\EndWhile
\end{algorithmic}
\end{algorithm}
\end{minipage}

\section{Experiments}
% The proposed method is implemented on two existing distillation approaches: Progressive Distillation and SFD.
\subsection{Experiment Settings}
\noindent
\textbf{Experimental Settings for PD}. The experiments utilize the pretrained model and weights from \cite{ho2020denoisingdiffusionprobabilisticmodels} as described in the original DDPM paper. The CIFAR-10 dataset, which contains $60,000$  $32 \times 32$ colour images across $10$ classes, is used for all experiments. Following standard protocols, we employ $\epsilon$-parameterization and a linear noise schedule to ensure consistency with recent literature. The teacher model initially performs $1024$ steps, which are distilled to $128$ steps. As the original authors set $K=2$, the same parameterization is used in B-DENSE. Each distillation iteration involves $50$k parameter updates, a batch size of $128$, and the AdamW optimizer with a learning rate of $2 \times 10^{-4}$
. The model's performance proved robust to variations in branch weights; thus, a uniform weighting strategy was used. Distillation experiments are conducted with NVIDIA L4 GPU and the time taken to distill a model is $\sim$ 4-5 hours.

\noindent
\textbf{Experimental Settings for SFD}. Pre-trained diffusion models provided by EDM~\cite{karras2022edm} were used. We evaluate results on CIFAR10 32$\times$32~\cite{krizhevsky2009learning} and ImageNet 64$\times$64~\cite{russakovsky2015ImageNet}. All training hyperparameters were set to the same values as described by the original authors. Consequently, $K$ was set to $4$. The values of $\{\lambda_i\}_{i=0}^{k-1}$ were set to $[0.017, 0.056, 0.191, 0.651]$ for CIFAR10 and $[0.014, 0.056, 0.223, 0.892]$ for ImageNet. Distillation on NFE 2 was done using NVIDIA A100 GPU and the time taken to distill on CIFAR-10 was $\sim$ 44 minutes and ImageNet was $\sim$ 3 hours. 

\begin{table}[h]
    \centering
    \caption{Comparison of the FID of both frameworks on CIFAR-10 with various NFE}
    \vspace{2pt}
    \label{Table1}

    \begin{tabular}{lcccccccc}
    \toprule
        \textbf{Methodology} & \textbf{512} & \textbf{256} & \textbf{128}  \\
        \midrule
        ProgressiveDistillation (Baseline) & 11.96 & 21.52 & 39.66 \\
        \textbf{B-DENSE (Ours)} & \textbf{8.92} & \textbf{12.04} & \textbf{20.81}   \\
        \bottomrule
    \end{tabular}
    
\end{table}

\subsection{Results}
\begin{figure}[h]
    \centering
    \includegraphics[width=0.45\linewidth]{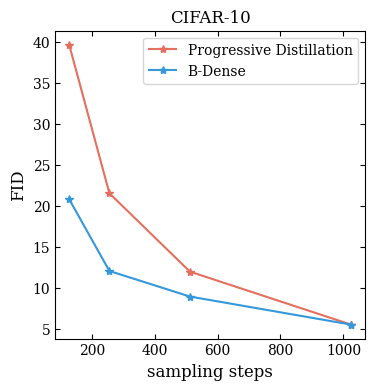}
    \caption{Results on calculating FID for 50k images}
    \label{fig:fid_50k}
\end{figure}

\textbf{Results on PD:} \autoref{Table1} shows that our technique achieves better performance than the baseline framework. 
\autoref{fig:fid_50k} shows that, as distillation progresses, the FID between the two models diverges. 
This indicates that our method provides richer supervision by leveraging the discarded intermediate timestep at each iteration, thereby reducing discretization error in the sampling trajectory. Our method achieves this with almost no training time increment.

\vspace{5pt}

\vspace{5pt}

\textbf{Results on SFD:} The integration of B-DENSE into the SFD framework yields consistent improvements in image fidelity across different sampling budgets. By providing dense trajectory supervision,  the model bridges the gap between the teacher’s continuous path and the student’s discretized steps.

As shown in \ref{tab:Table 2}, B-DENSE significantly improves performance in the ultra-low-step regime (NFE $2$) on ImageNet. We speculate that the suboptimal performance for B-DENSE for NFE $\geq 3$ on ImageNet is attributed to the fact that the branch weights were chosen according to the CIFAR 10 dataset, as discussed in \autoref{app:optimal_weights}. Our framework achieves an FID of $4.40$, a notable improvement over the baseline's $4.53$, proving that intermediate branch supervision helps the model maintain structural integrity even with minimal steps. Across all experiments, we observed virtually the same wall clock time and memory usage with and without B-DENSE.
\vspace{5pt}

\begin{table}[h]
    \centering
    % --- First Table (Left) ---
    \begin{minipage}[t]{0.48\textwidth}
        \centering
        \caption{FID comparison on CIFAR-10 }
        \label{tab:Table 2}
        \resizebox{\textwidth}{!}{% Optional: resizes table to fit width
        \begin{tabular}{lcccc}
            \toprule
            \textbf{Methodology} & \textbf{NFE 2} & \textbf{NFE 3} & \textbf{NFE 4} & \textbf{NFE 5} \\
            \midrule
            SFD (Baseline) & 4.53 & 3.58 & 3.24 & 3.06 \\
            \textbf{B-DENSE (Ours)} & \textbf{4.40} & \textbf{3.52} & \textbf{3.21} & \textbf{3.01} \\
            \bottomrule
        \end{tabular}
        }
    \end{minipage}
    \hfill % Adds flexible space between the tables
    % --- Second Table (Right) ---
    \begin{minipage}[t]{0.48\textwidth}
        \centering
        \caption{FID comparison on ImageNet 64x64 }
        \label{tab:Table 3}
        \resizebox{\textwidth}{!}{% Optional: resizes table to fit width
        \begin{tabular}{lcccc}
            \toprule
            \textbf{Methodology} & \textbf{NFE 2} & \textbf{NFE 3} & \textbf{NFE 4} & \textbf{NFE 5} \\
            \midrule
            SFD (Baseline) & 10.25 & \textbf{6.35} & \textbf{4.99} & \textbf{4.33} \\
            \textbf{B-DENSE (Ours)} & \textbf{9.57} & 6.54 & 5.97 & 5.91 \\
            \bottomrule
        \end{tabular}
        }
    \end{minipage}
\end{table}

\vspace{5pt}

The scalability of B-DENSE is further evaluated on the ImageNet dataset. As the image distribution becomes more complex and diverse, dense supervision from auxiliary branches plays a key role in maintaining fine-grained visual structure and stable denoising behavior. Results are summarized in \ref{tab:Table 3}.

The results indicate a clear trend: by supervising the student on the trajectory path itself rather than just the final endpoint, B-DENSE achieves a more robust approximation of the Probability Flow ODE, reducing the cumulative error that we usually encounter in low-step regimes.
\vspace{5pt}

\begin{figure}[H]
    \centering
    % First Image
    \includegraphics[width=0.45\linewidth]{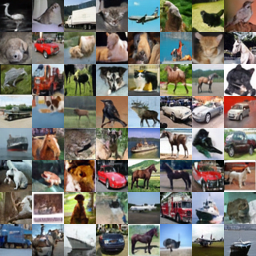}
 \hspace{35pt}% Adds horizontal space between the two images
    % Second Image
    \includegraphics[width=0.45\linewidth]{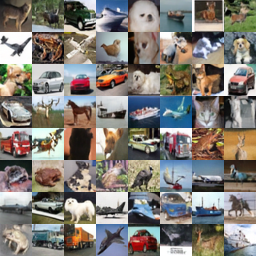}
    
    \caption{Comparison of results of B-DENSE (right) and SFD (left)}
    \label{fig:combined_results}
\end{figure}

\vspace{5pt}
\section{Discussion}
The empirical success of B-DENSE across multiple frameworks (Progressive Distillation and SFD) suggests that the missing link in current distillation research is not necessarily the capacity of the student model, but the density of the supervision signal.

\textbf{Mitigating Discretization Error.} Standard distillation forces the student to learn a shortcut between two distant points on the PF-ODE. In a high-curvature vector field, this linear approximation leads to discretization error, which compounds over sampling steps. By enforcing alignment at $K$ intermediate points, B-DENSE effectively constrains the student to follow the teacher's curved trajectory. This reduces the variance of the update even when the number of function evaluations is small.

\textbf{Computational Efficiency and Scalability.} A primary advantage of B-DENSE is its free lunch nature. Since the branched outputs share the entire backbone of the U-NET and are only active during training, the method provides a significant reduction in the discretization errors with near-zero impact on forward and backward pass latency. This makes it particularly attractive for scaling distillation to high-resolution models like Stable Diffusion, where training costs are otherwise prohibitive.

\textbf{Limitations and future work.} While B-DENSE significantly improves sampling quality, it remains tethered to the teacher’s trajectory quality; any inherent artifacts in the teacher’s sampling path are faithfully replicated by the student. Furthermore, the current dependence on framework-specific branch weights  provides a natural entry point for developing more adaptive, self-tuning schedules. Future work will focus on transforming these branch weights into learnable parameters, allowing the model to dynamically balance structural alignment and perceptual refinement. We also aim to validate this framework on Latent Diffusion Models and scale it toward Video and 3D generation, where dense trajectory consistency is paramount.

\bibliography{iclr2026_delta}

@misc{ho2020denoisingdiffusionprobabilisticmodels,
      title={Denoising Diffusion Probabilistic Models}, 
      author={Jonathan Ho and Ajay Jain and Pieter Abbeel},
      year={2020},
      eprint={2006.11239},
      archivePrefix={arXiv},
      primaryClass={cs.LG},
      url={https://arxiv.org/abs/2006.11239}, 
}

@misc{salimans2022progressivedistillationfastsampling,
      title={Progressive Distillation for Fast Sampling of Diffusion Models}, 
      author={Tim Salimans and Jonathan Ho},
      year={2022},
      eprint={2202.00512},
      archivePrefix={arXiv},
      primaryClass={cs.LG},
      url={https://arxiv.org/abs/2202.00512}, 
}

@misc{song2021scorebasedgenerativemodelingstochastic,
      title={Score-Based Generative Modeling through Stochastic Differential Equations}, 
      author={Yang Song and Jascha Sohl-Dickstein and Diederik P. Kingma and Abhishek Kumar and Stefano Ermon and Ben Poole},
      year={2021},
      eprint={2011.13456},
      archivePrefix={arXiv},
      primaryClass={cs.LG},
      url={https://arxiv.org/abs/2011.13456}, 
}

@inproceedings{
zhou2024simple,
title={Simple and Fast Distillation of Diffusion Models},
author={Zhenyu Zhou and Defang Chen and Can Wang and Chun Chen and Siwei Lyu},
booktitle={The Thirty-eighth Annual Conference on Neural Information Processing Systems},
year={2024},
url={https://openreview.net/forum?id=Ao0FiZqrXa}
}

@misc{song2023consistencymodels,
      title={Consistency Models}, 
      author={Yang Song and Prafulla Dhariwal and Mark Chen and Ilya Sutskever},
      year={2023},
      eprint={2303.01469},
      archivePrefix={arXiv},
      primaryClass={cs.LG},
      url={https://arxiv.org/abs/2303.01469}, 
}

@misc{kim2024consistencytrajectorymodelslearning,
      title={Consistency Trajectory Models: Learning Probability Flow ODE Trajectory of Diffusion}, 
      author={Dongjun Kim and Chieh-Hsin Lai and Wei-Hsiang Liao and Naoki Murata and Yuhta Takida and Toshimitsu Uesaka and Yutong He and Yuki Mitsufuji and Stefano Ermon},
      year={2024},
      eprint={2310.02279},
      archivePrefix={arXiv},
      primaryClass={cs.LG},
      url={https://arxiv.org/abs/2310.02279}, 
}

@misc{zhu2025distillingparallelgradientsfast,
      title={Distilling Parallel Gradients for Fast ODE Solvers of Diffusion Models}, 
      author={Beier Zhu and Ruoyu Wang and Tong Zhao and Hanwang Zhang and Chi Zhang},
      year={2025},
      eprint={2507.14797},
      archivePrefix={arXiv},
      primaryClass={cs.CV},
      url={https://arxiv.org/abs/2507.14797}, 
}

@inproceedings{sohl2015deep,
  title={Deep unsupervised learning using nonequilibrium thermodynamics},
  author={Sohl-Dickstein, Jascha and Weiss, Eric and Maheswaranathan, Niru and Ganguli, Surya},
  booktitle={International conference on machine learning},
  pages={2256--2265},
  year={2015},
  organization={PMLR}
}

@inproceedings{song2019ncsn,
	author    = {Yang Song and Stefano Ermon},
	title     = {Generative Modeling by Estimating Gradients of the Data Distribution},
	booktitle = {Advances in Neural Information Processing Systems},
	year      = {2019},
}

@inproceedings{ho2020ddpm,
	author={Ho, Jonathan and Jain, Ajay and Abbeel, Pieter},
	title={Denoising Diffusion Probabilistic Models},
	booktitle={Advances in Neural Information Processing Systems},
	year={2020}
}

@article{kingma2013auto,
  title={Auto-encoding variational bayes},
  author={Kingma, Diederik P and Welling, Max},
  journal={arXiv preprint arXiv:1312.6114},
  year={2013}
}

@inproceedings{karras2022edm,
	author    = {Tero Karras and Miika Aittala and Timo Aila and Samuli Laine},
	title     = {Elucidating the Design Space of Diffusion-Based Generative Models},
	booktitle={Advances in Neural Information Processing Systems},
	year      = {2022}
}

@inproceedings{salimans2022progressive,
	title={Progressive Distillation for Fast Sampling of Diffusion Models},
	author={Tim Salimans and Jonathan Ho},
	booktitle={International Conference on Learning Representations},
	year={2022},
}

@article{krizhevsky2009learning,
	title={Learning multiple layers of features from tiny images},
	author={Krizhevsky, Alex and Hinton, Geoffrey},
	journal={Technical Report},
	year={2009}
}

@article{russakovsky2015ImageNet,
	author    = {Olga Russakovsky and
	Jia Deng and
	Hao Su and
	Jonathan Krause and
	Sanjeev Satheesh and
	Sean Ma and
	Zhiheng Huang and
	Andrej Karpathy and
	Aditya Khosla and
	Michael S. Bernstein and
	Alexander C. Berg and
	Fei{-}Fei Li},
	title     = {ImageNet Large Scale Visual Recognition Challenge},
	journal   = {International Journal of Computer Vision},
	volume    = {115},
	number    = {3},
	pages     = {211--252},
	year      = {2015},
}

@inproceedings{heusel2017gans,
  author        = {Heusel, Martin and Ramsauer, Hubert and Unterthiner, Thomas and Nessler, Bernhard and Hochreiter, Sepp},
  title         = {{GANs} trained by a two time-scale update rule converge to a local {Nash} equilibrium},
  year          = {2017},
  booktitle     = {Advances in Neural Information Processing Systems},
  pages         = {6626--6637},
}

@article{kim2023consistency,
  title={Consistency trajectory models: Learning probability flow ode trajectory of diffusion},
  author={Kim, Dongjun and Lai, Chieh-Hsin and Liao, Wei-Hsiang and Murata, Naoki and Takida, Yuhta and Uesaka, Toshimitsu and He, Yutong and Mitsufuji, Yuki and Ermon, Stefano},
  journal={arXiv preprint arXiv:2310.02279},
  year={2023}
}

@InProceedings{Xiang_2025_CVPR,
    author    = {Xiang, Qianlong and Zhang, Miao and Shang, Yuzhang and Wu, Jianlong and Yan, Yan and Nie, Liqiang},
    title     = {DKDM: Data-Free Knowledge Distillation for Diffusion Models with Any Architecture},
    booktitle = {Proceedings of the IEEE/CVF Conference on Computer Vision and Pattern Recognition (CVPR)},
    month     = {June},
    year      = {2025},
    pages     = {2955-2965}
}
\bibliographystyle{iclr2026_delta}

\appendix
\section{Appendix}
\label{appendix}
\subsection{Choosing Optimal Branch Weights}
\label{app:optimal_weights}

To determine the optimal contribution of each branch to the total loss, we treated the weighting coefficients $\{\lambda_i\}_{i=0}^{K-1}$ as hyperparameters. Conducting a grid search over four independent continuous variables is computationally expensive and prone to instability. Instead, we assumed that the supervision signal should monotonically increase as the trajectory approaches the final clean data state. This reflects the intuition that errors in the final steps of generation have a more direct impact on perceptual quality than errors in the noisy, early stages, due to the lower Signal-to-Noise-Ratio (SNR).

Based on this assumption, we hypothesized that the optimal weight schedule follows a geometric progression. We parametrized the weights using a linear function in the log-space:

\begin{equation}
    \lambda_i = \exp(a \cdot i + b)
\end{equation}

where $i \in \{0, \dots, K-1\}$. This formulation reduces the search space to two parameters: $a$, which controls the rate of growth (slope), and $b$, which determines the initial magnitude (intercept).

\paragraph{Search Strategy and Constraints.}
Performing a comprehensive hyperparameter search for Progressive Distillation (PD) is computationally prohibitive, as the pipeline requires multiple sequential training sessions to iteratively halve the step count. To overcome this limitation, we conducted the search using the Simple and Fast Distillation (SFD) framework. We specifically targeted the CIFAR-10 dataset in the NFE $2$ (Number of Function Evaluations) regime. This configuration provided the fastest training and evaluation loop, enabling us to efficiently explore the parameter space. We reasoned that weights optimized for the most constrained sampling regime (2 steps) on a standard dataset would transfer effectively to other resolutions and step counts.

We utilized the Optuna framework to search for $a$ and $b$, ensuring numerical stability by constraining $b \in [-7, -2]$. We then validated the transferability of the optimal weights found on CIFAR-10 by applying them to ImageNet $64\times64$.

\paragraph{Results.} The search identified a distinct 'valley' of optimal configurations where the weights follow a steep exponential curve (high $a$). This confirms our initial assumption: assigning significantly higher importance to the later branches (closer to $x_0$) yields better convergence. Table \ref{tab:table 4} presents the top-5 configurations. While B-DENSE provides the most significant gains at NFE 2, where discretization error is most acute, the dense constraints serve as a powerful regularizer that maintains structural integrity across the low-step regime. The results for B-DENSE on Imagenet for NFE$\ge 3$ being suboptimal can be largely attributed to lack of hyperparameter search under compute limitations.

\begin{figure}[!t]
    \centering
    \includegraphics[width=0.6\textwidth]{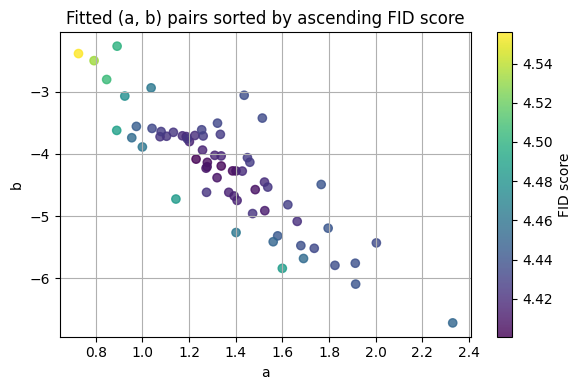} 
    \caption{Fitted $(a, b)$ parameter pairs sorted by ascending FID score. The color scale indicates the FID score (lower is better). The clustering of high-performing configurations (dark purple) demonstrates a strong preference for a specific relationship between the growth rate $a$ and intercept $b$.}
    \label{fig:fitted_ab}
\end{figure}

\begin{table}[t]
    \centering
    \caption{Top-5 weight configurations found via Optuna parameter search. The weights follow the form $\lambda_i = \exp(a \cdot i + b)$. We report FID scores at NFE 2 for both ImageNet $64\times64$ and CIFAR-10 $32\times32$. All configurations outperform the SFD baseline.}
    \label{tab:table 4}
    \begin{tabular}{@{}lccc@{}}
        \toprule
         & & \multicolumn{2}{c}{\textbf{FID (NFE 2)}} \\
        \cmidrule(l){3-4}
        \textbf{a} & \textbf{b} & \textbf{ImageNet 64} & \textbf{CIFAR-10} \\
        \midrule
        1.386 & -4.274 & \textbf{9.57} & \textbf{4.40} \\
        1.337 & -4.195 & 9.72 & 4.40 \\
        1.230 & -4.085 & 9.84 & 4.40 \\
        1.278 & -4.137 & 9.72 & 4.41 \\
        1.277 & -4.207 & 9.64 & 4.41 \\
        \midrule
        SFD baseline &  & 10.57 & 4.53 \\
        \bottomrule
    \end{tabular}
\end{table}
\end{document}